\pdfoutput=1
\documentclass[11pt,a4paper]{article}

\usepackage[T2A]{fontenc}
\usepackage[utf8]{inputenc}
\usepackage[english]{babel}

\usepackage{amsmath,amssymb}
\usepackage{booktabs}
\usepackage{multirow}
\usepackage{array}
\usepackage{placeins}

\usepackage{graphicx}
\usepackage{float}
\usepackage[protrusion=true,expansion=true,disable={footnote}]{microtype}
\usepackage{enumitem}
\usepackage{geometry}
\usepackage{flafter}

\usepackage[numbers,sort&compress]{natbib}
\usepackage{hyperref}
\hypersetup{
    colorlinks=true,
    linkcolor=blue,
    citecolor=blue,
    urlcolor=blue
}
\usepackage{setspace}
\usepackage{caption}
\usepackage{subcaption}

\title{\textbf{FINESSE-Bench: A Hierarchical Benchmark Suite for Financial Domain Knowledge and Technical Analysis in Large Language Models}}

\author{
Dmitry Stanishevskii \\
Lime FinTech \and
Nini Kamkia \\
Lime FinTech \and
Alexey Khoroshilov \\
Lime FinTech \and
Dmitry Zmitrovich \\
Lime FinTech \and
Denis Kokosinskii \\
Lime FinTech \and
Zhirayr Hayrapetyan \\
Lime FinTech \and
Andrei Kalmykov \\
Lime FinTech
}

\date{}

\begin{document}

\maketitle

\begin{abstract}
Large language models (LLMs) are increasingly being applied to financial analysis, reporting, investment decision support, risk management, compliance, and professional training. However, robust evaluation of their domain competence in finance remains incomplete. Widely used open benchmarks such as FinQA, ConvFinQA, and TAT-QA have played an important role in advancing financial question answering and numerical reasoning, but they focus primarily on question answering over financial reports and do not provide an explicit hierarchy of professional difficulty \cite{chen2021finqa, chen2022convfinqa, zhu2021tatqa}. Broader resources, including FinanceBench, PIXIU, FinBen, and FLaME, expand the coverage of financial tasks, yet the problem of evaluating the transition from foundational knowledge to expert-level financial reasoning remains open \cite{islam2023financebench, xie2023pixiu, xie2024finben, matlin2025flame}.

In this work, we present \textbf{FINESSE-Bench}, a suite of eight specialized benchmarks comprising 3,993 questions for hierarchical evaluation of financial competencies in LLMs. FINESSE-Bench combines exam-oriented datasets inspired by professional certifications (CFA-like Levels~1--3, CMT-like Level~2, and CFTe-like Level~1), applied trading task collections, and a Russian-language olympiad benchmark. This design enables evaluation of domain breadth, performance degradation as difficulty increases, the ability to solve computational tasks, and model behavior in specialized financial domains.

We also describe a unified evaluation protocol covering multiple-choice questions, numerical answers, and short open-ended responses, together with an automated scoring scheme for free-form answers based on the LLM-as-judge paradigm \cite{zheng2023judge, li2024arenahard}. FINESSE-Bench is intended both as a complement to existing open financial benchmarks and as a tool for more substantive evaluation of professionally relevant financial competencies in large language models.
\end{abstract}

\section{Introduction}

Large language models (LLMs) have advanced substantially in text understanding, reasoning, and structured generation, which has stimulated their adoption across the financial industry, including financial analysis, reporting, investment research, risk management, compliance, and professional training. However, deployment in high-stakes settings requires reliable evaluation of models' domain competencies, spanning financial reporting, corporate finance, portfolio management, derivatives, and technical analysis.

Over the past several years, important open benchmarks have been introduced for evaluating models in finance. FinQA, ConvFinQA, and TAT-QA laid the foundation for financial question answering and numerical reasoning over financial documents and hybrid table-text data \cite{chen2021finqa, chen2022convfinqa, zhu2021tatqa}. More recent work has broadened task coverage: FinanceBench introduced a large open benchmark for financial question answering over public-company documents \cite{islam2023financebench}; PIXIU, FinBen, and FLaME extended this line toward broader evaluation of language models and financial NLP tasks \cite{xie2023pixiu, xie2024finben, matlin2025flame}.

Despite the value of these resources, at least two limitations remain. First, a substantial portion of existing benchmarks is concentrated on question answering over financial reports, information extraction, or financial NLP, while several practically important areas---such as technical analysis, derivatives trading, and portfolio management in scenario-based settings---remain underrepresented. Second, most open financial benchmarks lack an explicit difficulty hierarchy that would allow one to measure how model behavior changes when moving from basic financial knowledge to expert-level tasks requiring multi-step analysis and synthesis.

Additional motivation for more challenging and more diagnostic benchmarks comes from recent literature on financial numerical reasoning. In particular, FinanceReasoning emphasizes that financial benchmarks should be evaluated not only in terms of popularity but also in terms of fidelity, difficulty, and completeness of financial concept coverage \cite{tang2025financereasoning}. In parallel, the development of specialized models such as Fin-R1 shows that strong results on standard public datasets do not necessarily provide a comprehensive characterization of model behavior across a broader spectrum of professional financial tasks \cite{finr12025}.

In this work, we present \textbf{FINESSE-Bench}, a hierarchical benchmark suite for evaluating financial competencies in LLMs. FINESSE-Bench includes eight datasets with a total of 3,993 questions and combines two key principles. The first is a \textit{difficulty hierarchy}: part of the suite is inspired by the structure of professional certifications and allows one to measure the transition from foundational to advanced and expert-level competence. The second is \textit{domain specialization}: in addition to classical finance disciplines, the suite covers technical analysis, applied derivatives trading, and Russian-language olympiad-style problems.

Beyond dataset construction, we describe a unified evaluation protocol applicable to heterogeneous task types: multiple-choice questions, numerical answers, short free-form answers, and case-linked questions. For tasks where exact matching is insufficient, we use an LLM-as-judge scheme grounded in modern approaches to automated evaluation of open-ended responses \cite{zheng2023judge, li2024arenahard}. Full results, extended tables, and supplementary materials for the benchmark suite are released in the project repository: \url{https://github.com/LimexAILab/FINESSE-Bench}.

\paragraph{Contributions.}
Our work makes the following contributions:
\begin{enumerate}[leftmargin=*,itemsep=2pt]
    \item We introduce \textbf{FINESSE-Bench}, a suite of eight specialized financial benchmarks comprising 3,993 questions.
    \item We propose a \textbf{hierarchical evaluation design} that enables measurement of model performance degradation when moving from basic to advanced and expert-level financial difficulty.
    \item We broaden \textbf{domain coverage} beyond standard financial-report QA by including technical analysis, derivatives trading, and a Russian-language olympiad block.
    \item We describe a \textbf{unified evaluation protocol} for heterogeneous financial tasks, combining fixed prompting templates, deterministic inference settings where applicable, and judge-model-based scoring for open-ended answers.
    \item We release the datasets for \textbf{non-commercial research use} and discuss limitations related to data provenance, possible contamination, and licensing.
\end{enumerate}

\section{Related Work}
\label{sec:related_work}

\subsection{Financial Benchmarks for LLMs}

The development of financial benchmarks for language models has accelerated in recent years. One of the earliest important directions was the creation of datasets for financial question answering and numerical reasoning. FinQA introduced an expert-annotated dataset of questions and answers over financial reports with executable reasoning programs \cite{chen2021finqa}. ConvFinQA extended this setup to conversational financial QA, where longer chains of numerical reasoning are required \cite{chen2022convfinqa}. TAT-QA proposed a hybrid format combining tabular and textual sources in financial question answering tasks \cite{zhu2021tatqa}.

Later work introduced broader resources. FinanceBench proposed a large open benchmark for financial question answering over public-company documents \cite{islam2023financebench}. PIXIU presented a financial ecosystem including instruction data, a model, and a benchmark component covering multiple types of financial tasks \cite{xie2023pixiu}. FinBen expanded the line of comprehensive evaluation by aggregating dozens of datasets and task types across multiple financial domains \cite{xie2024finben}. FLaME continued this direction by providing a broader platform for evaluating financial language models \cite{matlin2025flame}.

These efforts have advanced the field substantially, but they do not fully address the problem of \textit{hierarchical} evaluation of professional financial competence. In particular, existing resources often lack the simultaneous combination of three properties: explicit difficulty gradation, grounding in professionally recognizable levels of expertise, and broader coverage of applied financial domains.

\subsection{Evaluation of Free-Form Answers and the LLM-as-Judge Paradigm}

As benchmark tasks move from exact-answer matching toward more open-ended response formats, automated evaluation becomes more difficult. Zheng et al.\ showed that strong language models can serve as judges for scalable evaluation of open-ended responses and systematized the limitations of this approach, including bias and prompt sensitivity \cite{zheng2023judge}. Subsequent work, including Arena-Hard and BenchBuilder, demonstrated that LLM-based evaluation can be useful not only for ranking models but also for constructing more discriminative benchmarks \cite{li2024arenahard}.

In our work, model-as-judge evaluation is used as a practical and reproducible mechanism for unified assessment of heterogeneous open-form financial tasks. At the same time, we do not treat automatic judge-based scoring as a complete substitute for expert annotation, but rather as a scalable compromise for large-scale benchmark evaluation.

\subsection{Difficulty, Fidelity, and Robustness of Financial Benchmarks}

Recent work such as FinanceReasoning emphasizes that financial benchmarks should be analyzed not only in terms of size, but also in terms of fidelity, coverage completeness, and genuine task difficulty \cite{tang2025financereasoning}. In particular, the authors revise and update parts of existing financial numerical reasoning benchmarks, underscoring the importance of benchmark design quality as an independent research topic.

At the same time, the development of specialized models, including Fin-R1, suggests that strong results on standard public financial benchmark datasets are useful but do not necessarily reflect robust professional competence across a broader range of financial scenarios \cite{finr12025}. This creates further motivation for benchmarks that evaluate not only accuracy in a narrow format, but also breadth of domain coverage, skill transfer, and changes in performance across difficulty levels.

\subsection{Professionally Oriented and Domain-Specialized Benchmarks}

Using exam-style and professionally oriented tasks is a natural way to evaluate domain competence in applied fields. In finance, this approach is particularly appropriate because a substantial portion of professional knowledge is already structured in the form of certifications and applied work scenarios.

FINESSE-Bench follows precisely this logic: we construct a suite of complementary benchmarks, some of which reflect progression from foundational preparation to expert-level tasks, while others target practice-oriented domains that are underrepresented in existing open resources.

\section{FINESSE-Bench: Design Principles}
\label{sec:design}

In designing FINESSE-Bench, we started from the premise that financial competence in LLMs is not a one-dimensional quantity. The same model may answer basic financial reporting questions confidently while performing noticeably worse on portfolio construction, technical analysis, or derivatives trading tasks. A financial benchmark suite should therefore evaluate not only average accuracy, but also the \textit{structure of model errors} across task types and difficulty levels.

\paragraph{Realism.}
We aimed for questions that reflect skills relevant to real financial practice and professional training: interpretation of financial statements, company valuation, risk management, investment decision-making, use of technical indicators, and option-strategy calculations.

\paragraph{Difficulty hierarchy.}
A central principle of FINESSE-Bench is explicit difficulty gradation. Inspired by multi-level professional certifications, we include task sets corresponding to foundational, intermediate, and expert levels. This makes it possible to measure how well a model transfers basic knowledge to more complex scenario-based and multi-step tasks.

\paragraph{Domain breadth.}
Existing open benchmark resources in finance are particularly strong in question answering over financial reporting and financial NLP tasks \cite{chen2021finqa, chen2022convfinqa, zhu2021tatqa, islam2023financebench, xie2024finben}. We complement this line with datasets on technical analysis, derivatives trading, and Russian-language olympiad problems in order to broaden the range of competencies that can be diagnosed.

\paragraph{Format diversity.}
FINESSE-Bench includes multiple-choice questions, numerical answers, short free-form responses, and linked case-based questions. Such diversity makes it more difficult to optimize narrowly for a single evaluation format while also bringing the benchmark closer to real educational and professional scenarios.

\paragraph{Multilinguality.}
Although most open financial benchmarks are in English, practical applications of LLMs in finance are often multilingual. For this reason, FINESSE-Bench includes the Russian-language block VLigaBench-ru, enabling evaluation of model behavior beyond English.

\paragraph{Verifiability.}
All questions are paired with verifiable answers, and some tasks also include short justifications or calculation templates. This facilitates automated scoring and error analysis and makes the benchmark suite more suitable for reproducible comparison.

\section{Dataset Description}
\label{sec:datasets}

FINESSE-Bench consists of eight specialized datasets comprising a total of 3,993 questions. Below, we briefly describe their purpose and role in the overall hierarchy of the benchmark suite.

\subsection{CFA-like Level~1}
CFA-like Level~1\footnote{\url{https://www.cfainstitute.org/programs/cfa-program}} targets foundational finance disciplines: ethics, quantitative methods, economics, financial reporting, corporate finance, and investment fundamentals. The benchmark includes 1,069 questions, predominantly in multiple-choice format. Its purpose is to measure basic financial literacy and applied competence.

\subsection{CFA-like Level~2}
CFA-like Level~2\footnote{\url{https://www.cfainstitute.org/programs/cfa-program}} focuses on more complex application scenarios. It contains 293 questions organized into linked item sets, where several interrelated questions rely on a common case. Multi-step calculations, advanced financial statement analysis, valuation, fixed income, and derivatives all play an important role here.

\subsection{CFA-like Level~3}
CFA-like Level~3\footnote{\url{https://www.cfainstitute.org/programs/cfa-program}} targets expert-level tasks in portfolio management, private wealth planning, risk management, and complex ethical case analysis. The benchmark contains 318 questions and is intended to measure expert competence requiring strategic thinking and synthesis across multiple areas of finance.

\subsection{CMT-like Level~2}
CMT-like Level~2\footnote{\url{https://cmtassociation.org/}} contains 251 questions on technical analysis and market statistics, including technical analysis theory, chart patterns, indicators, volume, open interest, trading-system testing, and risk management. This dataset diagnoses more applied skills related to working with market signals.

\subsection{CFTe-like Level~1}
CFTe-like Level~1\footnote{\url{https://www.ifta.org/certified-financial-technician-cfte-}} contains 781 questions on basic concepts in technical analysis: chart types, trends, support and resistance levels, basic patterns, moving averages, and momentum indicators. Within the full collection, it serves as the foundational technical-analysis block.

\subsection{VLigaBench-ru}
VLigaBench-ru is a Russian-language olympiad-style dataset of 324 problems in microeconomics, macroeconomics, financial mathematics, and game theory. Unlike typical financial QA tasks, this dataset places stronger emphasis on reasoning, calculation, and careful handling of Russian-language problem statements.

\subsection{Trading\_TA}
Trading\_TA contains 413 applied technical-analysis tasks in a trading context: pattern recognition, momentum and mean-reversion strategies, entry and exit rules, stop management, backtesting on historical data, and multi-timeframe analysis. This block is intended to assess more practice-oriented competence.

\subsection{Trading\_derivatives}
Trading\_derivatives consists of 544 tasks on options, synthetic positions, put-call parity, arbitrage, Greeks, hedging, pricing, and futures strategies. This dataset is one of the most specialized and calculation-intensive components of FINESSE-Bench.

\subsection{Dataset Statistics}

\begin{table}[H]
\centering
\caption{Core statistics of the FINESSE-Bench datasets.}
\label{tab:dataset_stats}
\begin{tabular}{lcccc}
\toprule
\textbf{Dataset} & \textbf{Questions} & \textbf{Format} & \textbf{Avg. length, characters} & \textbf{Language} \\
\midrule
CFA-like Level~1      & 1069 & MCQ      & 348  & en \\
CFA-like Level~2      & 293  & MCQ      & 2965 & en \\
CFA-like Level~3      & 318  & MCQ      & 3951 & en \\
CMT-like Level~2      & 251  & MCQ      & 297  & en \\
CFTe-like Level~1     & 781  & MCQ      & 368  & en \\
VLigaBench-ru         & 324  & NAQ/SAQ  & 596  & ru \\
Trading\_TA           & 413  & NAQ/SAQ  & 280  & en \\
Trading\_derivatives  & 544  & NAQ/SAQ  & 292  & en \\
\midrule
\textbf{Total}        & \textbf{3993} & \textbf{mixed} & \textbf{833} & \textbf{en/ru} \\
\bottomrule
\end{tabular}
\end{table}

\paragraph{Format notation.}
MCQ denotes multiple-choice questions; NAQ denotes numerical-answer questions; SAQ denotes short-answer questions.

\subsection{Data Collection and Curation}

The questions were collected from publicly available internet sources, educational materials, training problems, publicly available exam-style explanations, and olympiad problems. After collection, the data underwent normalization of format, alignment of answer structure, and manual checking for basic correctness.

It is important to note that the provenance of individual questions was not fully documented during data accumulation. This creates limitations in terms of complete traceability and requires a cautious distribution policy. For this reason, the datasets are released under a non-commercial license, and a removal mechanism for disputed materials is provided through the project repository.

We also acknowledge the possibility that some questions may overlap with the training data of certain models, as well as potential biases arising from uneven topic and source coverage. These limitations are discussed further in Section~\ref{sec:practical_limitations}.

\section{Evaluation Protocol}
\label{sec:evaluation}

\subsection{Evaluated Models}

FINESSE-Bench is intended for comparison across a broad range of models: closed frontier models, open general-purpose models, specialized financial models, and reasoning-oriented models. The full list of models and exact inference configurations is available in the accompanying project repository.

Where applicable, models were evaluated in their reasoning (``thinking'') configurations. For readability, the main text and tables use normalized model names rather than full API or checkpoint identifiers; detailed model variants, exact inference settings, and complete results are provided in the project repository.

\subsection{Inference Settings}

Across all experiments, we use a unified fixed prompt template for each task type. To ensure comparability across models, prompts are provided without few-shot demonstrations. Wherever applicable, deterministic generation settings are used, including temperature $0.0$, unless constrained by API limitations or the recommended settings of a particular model.

For models that support controllable reasoning, the reasoning effort during scoring was set to medium.

For some models, inference is performed through a unified API provider, while for others it is run locally using inference tools. Importantly, each model configuration is fixed prior to evaluation and is not changed during the benchmark run.

\subsection{Scoring Scheme}

For all tasks, scoring is performed using a model-judge under the LLM-as-judge paradigm \cite{zheng2023judge, li2024arenahard}.

GPT-5.2 was used as the judge model. The judge model receives the question, the reference answer, and the tested model's response, and then assigns a binary correctness score. Our evaluation pipeline was initially adapted from the open-source \texttt{arena-hard-auto} framework and substantially extended for the FINESSE-Bench setting \cite{arenahardauto_repo}.

\subsection{Metrics}

The primary metric is accuracy:
$$
\mathrm{Accuracy}=\frac{1}{n}\sum_{i=1}^{n}s_i,\qquad s_i \in \{0,1\}.
$$

For each model on each benchmark, $95\%$ confidence intervals are computed using bootstrap. For aggregated benchmark groups, stratified bootstrap with weights proportional to dataset size is used.

In addition to per-dataset results, FINESSE-Bench supports group-level aggregation over three directions:
\begin{itemize}[leftmargin=*]
    \item \textbf{exam-like}: CFA-like Levels~1--3, CMT-like Level~2, VLigaBench-ru;
    \item \textbf{public benchmarks}: classical open financial benchmarks used for comparison against FINESSE-Bench;
    \item \textbf{trading/TA}: Trading\_derivatives, Trading\_TA, CFTe-like Level~1.
\end{itemize}

\subsection{Result Reporting Rules}

The main text reports only point estimates of accuracy. For all benchmark measurements, bootstrap confidence intervals and standard errors are also computed, but these are moved to the accompanying repository for compactness of the main presentation. Full results, including extended tables and additional configurations, are available in the project repository: \url{https://github.com/LimexAILab/FINESSE-Bench}.

\section{Main Results}
\label{sec:results}

In this section, we present the main results across several benchmark groups: classical open financial benchmarks, exam-oriented FINESSE-Bench tasks, and applied datasets focused on trading and technical analysis. This organization allows us to compare model behavior not only on widely used public financial evaluation sets, but also on more professionally oriented and domain-specialized tasks included in FINESSE-Bench.

Importantly, the tables below do not include every experiment we conducted, but rather a representative subset of the results. Our goal in the main text is to highlight the most informative patterns and cross-benchmark contrasts, while full results and additional configurations are provided in the accompanying repository and supplementary materials.

\begin{table}[t]
\centering
\small
\setlength{\tabcolsep}{4pt}
\caption{Results of Selected Models on Public Financial Benchmarks}
\label{tab:public_benchs_scores}
\begin{tabular}{lrrr}
\toprule
model &  FinQA &  ConvFinQA & TAT-QA \\
\midrule
Claude Sonnet 4.6 & 0.8274 & 0.9093 & 0.9586 \\
Claude Sonnet 3.7 & 0.7916 & 0.8905 & 0.9376 \\
Kimi K2.5 & 0.8178 & 0.9032 & 0.9538 \\
MiMo-V2-Flash & 0.7507 & 0.8481 & 0.9197 \\
GPT‑5.4& 0.8030 & 0.8925 & 0.9424 \\
GPT‑5.2& 0.8021 & 0.9187 & 0.9472 \\
MiniMax M2.5 & 0.8021 & 0.9039 & 0.9430 \\
MiniMax M2.1 & 0.7576 & 0.8676 & 0.8837 \\
Qwen3.5-Plus-02-15 & 0.7812 & 0.8978 & 0.9466 \\
Qwen3.5-Flash-02-23 & 0.7786 & 0.8837 & 0.9448 \\
Qwen3.5-397B-A17B & 0.7768 & 0.8884 & 0.9394 \\
Qwen3.5-122B-A17B & 0.7794 & 0.8878 & 0.9406 \\
Qwen3.5-35B-A3B & 0.7698 & 0.8851 & 0.9376 \\
Qwen3.5-27B & 0.7681 & 0.8898 & 0.9418 \\
Qwen3.5-9B & 0.7681 & 0.8757 & 0.9418 \\
GLM-5 & 0.7576 & 0.8898 & 0.9359 \\
GLM-4.7 & 0.7472 & 0.8911 & 0.9317 \\
Llama 4 Maverick & 0.7419 & 0.8394 & 0.9311 \\
Qwen3-235B-A22B-Thinking-2507  & 0.7751 & 0.8737 & 0.9251 \\
Qwen3-32B& 0.7847 & 0.8676 & 0.9305 \\
Qwen3-14B& 0.7786 & 0.8656 & 0.9335 \\
Qwen3-8B& 0.7568 & 0.8427 & 0.9221 \\
DeepSeek-V3.2 & 0.7629 & 0.8495 & 0.9287 \\
DeepSeek-R1-0528 & 0.7681 & 0.8642 & 0.9221 \\
Mistral-Small-3.2-24B-Instruct-2506 & 0.7629 & 0.8468 & 0.9293 \\
Fino1-8B & 0.7132 & 0.8024 & 0.8507 \\
Fin-R1 & 0.7742 & 0.8602 & 0.9107 \\
Fin-o1-8B & 0.7594 & 0.8454 & 0.9263 \\
GigaChat3-10B-A1.8B-bf16 & 0.7097 & 0.7856 & 0.8975 \\
YandexGPT Pro 5.1 & 0.7315 & 0.8522 & 0.9137 \\
T-pro-it-2.0 & 0.7777 & 0.8690 & 0.9317 \\
\bottomrule
\end{tabular}
\end{table}
\FloatBarrier

Table~\ref{tab:public_benchs_scores} presents results on classical open financial benchmarks. Overall, the top of the ranking on these benchmarks is relatively compressed: several strong models achieve similar accuracy values, and the gap among leading systems remains modest. In other words, while these benchmarks remain useful as a common reference point, they often provide only limited separation among the strongest contemporary models.

This pattern aligns with our original motivation for treating FINESSE-Bench not as a replacement for existing public resources, but as an additional instrument for more fine-grained evaluation of financial competence. In particular, it suggests that publicly established benchmarks are valuable for broad comparison, yet may be less informative when the goal is to diagnose more subtle differences in professionally relevant financial reasoning.

\begin{table}[t]
\centering
\scriptsize
\setlength{\tabcolsep}{1.2pt}
\caption{Results of Selected Models on Exam-Oriented FINESSE-Bench Benchmarks}
\label{tab:exam_like_scores}
\begin{tabular}{@{}lrrrrr@{}}
\toprule
model &
{\scriptsize\strut CFA-like Level~1} &
{\scriptsize\strut CFA-like Level~2} &
{\scriptsize\strut CFA-like Level~3} &
{\scriptsize\strut CMT-like Level~2} &
{\scriptsize\strut VLigaBench-ru} \\
\midrule
Claude Sonnet 4.6 & 0.8979 & 0.9181 & 0.8239 & 0.8924 & 0.8056 \\
Claude Sonnet 3.7 & 0.8419 & 0.8396 & 0.7893 & 0.8367 & 0.7901 \\
Kimi K2.5 & 0.8924 & 0.9181 & 0.8113 & 0.8805 & 0.8302 \\
MiMo-V2-Flash & 0.6642 & 0.4437 & 0.4780 & 0.6494 & 0.5278 \\
GPT‑5.4& 0.7315 & 0.6962 & 0.6509 & 0.8566 & 0.7932 \\
GPT‑5.2& 0.8736 & 0.8874 & 0.8019 & 0.8526 & 0.8488 \\
MiniMax M2.5 & 0.8148 & 0.8157 & 0.7170 & 0.8406 & 0.7901 \\
MiniMax M2.1 & 0.8092 & 0.7201 & 0.6981 & 0.8446 & 0.7068 \\
Qwen3.5-Plus-02-15 & 0.8896 & 0.9147 & 0.8208 & 0.8924 & 0.8488 \\
Qwen3.5-Flash-02-23 & 0.8662 & 0.8874 & 0.7453 & 0.8367 & 0.8117 \\
Qwen3.5-397B-A17B & 0.8756 & 0.8225 & 0.7201 & 0.8884 & 0.8488 \\
Qwen3.5-122B-A17B & 0.8803 & 0.8874 & 0.7673 & 0.8606 & 0.8364 \\
Qwen3.5-35B-A3B & 0.8634 & 0.8840 & 0.7516 & 0.8367 & 0.8056 \\
Qwen3.5-27B & 0.8587 & 0.8908 & 0.7642 & 0.8645 & 0.8117 \\
Qwen3.5-9B & 0.7774 & 0.7747 & 0.6855 & 0.7570 & 0.7840 \\
GLM-5 & 0.8859 & 0.8976 & 0.8396 & 0.8964 & 0.8210 \\
GLM-4.7 & 0.8765 & 0.9147 & 0.8019 & 0.8805 & 0.8241 \\
Llama 4 Maverick & 0.8241 & 0.7645 & 0.7170 & 0.8287 & 0.7068 \\
Qwen3-235B-A22B-Thinking-2507  & 0.8700 & 0.8635 & 0.7484 & 0.8446 & 0.7932 \\
Qwen3-32B& 0.8400 & 0.7679 & 0.7107 & 0.7968 & 0.7438 \\
Qwen3-14B& 0.8185 & 0.6212 & 0.6887 & 0.7809 & 0.6852 \\
Qwen3-8B& 0.7717 & 0.5666 & 0.5346 & 0.7450 & 0.6358 \\
DeepSeek-V3.2 & 0.7961 & 0.6587 & 0.5786 & 0.7530 & 0.7623 \\
DeepSeek-R1-0528 & 0.8587 & 0.8532 & 0.7736 & 0.8566 & 0.7253 \\
Mistral-Small-3.2-24B-Instruct-2506 & 0.6034 & 0.5051 & 0.5283 & 0.6653 & 0.5216 \\
Fino1-8B & 0.5229 & 0.4198 & 0.4245 & 0.5737 & 0.2778 \\
Fin-R1 & 0.6974 & 0.3891 & 0.4088 & 0.6375 & 0.4537 \\
Fin-o1-8B & 0.6501 & 0.4812 & 0.4686 & 0.6494 & 0.4815 \\
GigaChat3-10B-A1.8B-bf16 & 0.4911 & 0.3345 & 0.4151 & 0.6972 & 0.3920 \\
YandexGPT Pro 5.1 & 0.6791 & 0.5358 & 0.6006 & 0.7769 & 0.5556 \\
T-pro-it-2.0 & 0.8213 & 0.7679 & 0.6761 & 0.8008 & 0.7377 \\
\bottomrule
\end{tabular}
\end{table}
\FloatBarrier

Table~\ref{tab:exam_like_scores} reports results on the exam-oriented benchmarks. Here, differences between models are much more pronounced than on classical open benchmarks, leading to a clearer stratification of systems by capability. This already suggests that the exam-like benchmark group is more discriminative for evaluating professionally oriented financial knowledge and reasoning.

Moreover, within the \textit{exam-like} group, heterogeneity across difficulty levels is clearly visible: even strong models often exhibit performance drops when moving from CFA-like Level~1 to more difficult levels. This behavior is consistent with the original idea of hierarchical evaluation and indicates that the benchmark captures not only general financial familiarity, but also the ability of models to retain quality as task complexity increases.

\begin{table}[t]
\centering
\small
\setlength{\tabcolsep}{4pt}
\caption{Results of Selected Models on Trading and Technical Analysis Benchmarks}
\label{tab:ta_benchs_scores}
\begin{tabular}{lrrr}
\toprule
model & Trading\_derivatives & Trading\_TA & CFTe-like Level~1 \\
\midrule
Claude Sonnet 4.6 & 0.7849 & 0.8354 & 0.8464 \\
Claude Sonnet 3.7 & 0.7316 & 0.7797 & 0.8259 \\
Kimi K2.5 & 0.8658 & 0.8329 & 0.8476 \\
MiMo-V2-Flash & 0.5202 & 0.6368 & 0.7362 \\
GPT‑5.4& 0.7518 & 0.8087 & 0.8220 \\
GPT‑5.2& 0.8511 & 0.8354 & 0.8399 \\
MiniMax M2.5 & 0.7096 & 0.7627 & 0.7990 \\
MiniMax M2.1 & 0.6452 & 0.7579 & 0.8105 \\
Qwen3.5-Plus-02-15 & 0.8364 & 0.8232 & 0.8681 \\
Qwen3.5-Flash-02-23 & 0.7426 & 0.7724 & 0.8310 \\
Qwen3.5-397B-A17B & 0.8125 & 0.8184 & 0.8553 \\
Qwen3.5-122B-A17B & 0.8051 & 0.7821 & 0.8681 \\
Qwen3.5-35B-A3B & 0.7555 & 0.7869 & 0.8118 \\
Qwen3.5-27B & 0.7518 & 0.7966 & 0.8438 \\
Qwen3.5-9B & 0.6654 & 0.7385 & 0.7670 \\
GLM-5 & 0.8051 & 0.8257 & 0.8118 \\
GLM-4.7 & 0.7923 & 0.8160 & 0.8425 \\
Llama 4 Maverick & 0.6342 & 0.7458 & 0.8220 \\
Qwen3-235B-A22B-Thinking-2507  & 0.7684 & 0.7772 & 0.8399 \\
Qwen3-32B& 0.6801 & 0.7482 & 0.7951 \\
Qwen3-14B& 0.5901 & 0.7070 & 0.7772 \\
Qwen3-8B& 0.5551 & 0.6731 & 0.7580 \\
DeepSeek-V3.2 & 0.7739 & 0.7191 & 0.7721 \\
DeepSeek-R1-0528 & 0.7610 & 0.7651 & 0.8118 \\
Mistral-Small-3.2-24B-Instruct-2506 & 0.4136 & 0.6707 & 0.6684 \\
Fino1-8B & 0.1820 & 0.4407 & 0.6133 \\
Fin-R1 & 0.2923 & 0.5714 & 0.6543 \\
Fin-o1-8B & 0.3346 & 0.5763 & 0.6850 \\
GigaChat3-10B-A1.8B-bf16 & 0.2206 & 0.5617 & 0.6556 \\
YandexGPT Pro 5.1 & 0.4485 & 0.6683 & 0.7337 \\
T-pro-it-2.0 & 0.6636 & 0.7119 & 0.7759 \\
\bottomrule
\end{tabular}
\end{table}
\FloatBarrier

Table~\ref{tab:ta_benchs_scores} shows results on the applied benchmarks for technical analysis and derivatives trading. These datasets serve as an additional filter for models: in a number of cases, the ranking they induce does not match the ranking observed on public financial benchmarks. This discrepancy is important because it indicates that strong performance on standard public benchmarks does not automatically translate into equally strong performance in more specialized trading-related settings.

More broadly, these results suggest that domains such as technical analysis and derivatives contribute genuinely new diagnostic information. They capture aspects of financial competence that are only weakly reflected in more traditional benchmark formats and therefore help reveal strengths and weaknesses that might otherwise remain hidden in aggregated public-benchmark scores.

\begin{table}[t]
\centering
\small
\setlength{\tabcolsep}{4pt}
\caption{Aggregated Results by Benchmark Group}
\label{tab:agg_group_scores}
\begin{tabular}{lrrr}
\toprule
model & exam-like& public benchmarks & trading/TA \\
\midrule
Claude Sonnet 4.6 & 0.8763 & 0.9066 & 0.8245 \\
Claude Sonnet 3.7 & 0.8261 & 0.8824 & 0.7854 \\
Kimi K2.5 & 0.8740 & 0.9001 & 0.8498 \\
MiMo-V2-Flash & 0.5880 & 0.8499 & 0.6450 \\
GPT‑5.4& 0.7384 & 0.8880 & 0.7969 \\
GPT‑5.2& 0.8595 & 0.8987 & 0.8424 \\
MiniMax M2.5 & 0.8005 & 0.8919 & 0.7623 \\
MiniMax M2.1 & 0.7712 & 0.8445 & 0.7462 \\
Qwen3.5-Plus-02-15 & 0.8776 & 0.8856 & 0.8475 \\
Qwen3.5-Flash-02-23 & 0.8408 & 0.8794 & 0.7894 \\
Qwen3.5-397B-A17B & 0.8443 & 0.8784 & 0.8331 \\
Qwen3.5-122B-A17B & 0.8568 & 0.8794 & 0.8279 \\
Qwen3.5-35B-A3B & 0.8390 & 0.8747 & 0.7883 \\
Qwen3.5-27B & 0.8435 & 0.8775 & 0.8038 \\
Qwen3.5-9B & 0.7628 & 0.8726 & 0.7284 \\
GLM-5 & 0.8727 & 0.8724 & 0.8130 \\
GLM-4.7 & 0.8639 & 0.8685 & 0.8205 \\
Llama 4 Maverick & 0.7849 & 0.8489 & 0.7451 \\
Qwen3-235B-A22B-Thinking-2507  & 0.8381 & 0.8673 & 0.8027 \\
Qwen3-32B& 0.7938 & 0.8699 & 0.7479 \\
Qwen3-14B& 0.7512 & 0.8687 & 0.7019 \\
Qwen3-8B& 0.6891 & 0.8506 & 0.6744 \\
DeepSeek-V3.2 & 0.7379 & 0.8571 & 0.7601 \\
DeepSeek-R1-0528 & 0.8266 & 0.8610 & 0.7848 \\
Mistral-Small-3.2-24B-Instruct-2506 & 0.5751 & 0.8564 & 0.5892 \\
Fino1-8B & 0.4661 & 0.7974 & 0.4372 \\
Fin-R1 & 0.5650 & 0.8569 & 0.5212 \\
Fin-o1-8B & 0.5783 & 0.8538 & 0.5495 \\
GigaChat3-10B-A1.8B-bf16 & 0.4688 & 0.8087 & 0.4971 \\
YandexGPT Pro 5.1 & 0.6426 & 0.8438 & 0.6289 \\
T-pro-it-2.0 & 0.7796 & 0.8689 & 0.7255 \\
\bottomrule
\end{tabular}
\end{table}
\FloatBarrier

Aggregated results by benchmark group are reported in Table~\ref{tab:agg_group_scores}. This table most clearly illustrates the central thesis of the paper: strong results on classical open benchmarks do not always transfer to exam-oriented and applied professional tasks. In particular, for several models we observe a notable gap between performance on the \textit{public benchmarks} group and performance on the \textit{exam-like} and \textit{trading/TA} groups. This pattern supports the motivation for FINESSE-Bench as a benchmark suite designed not only to measure general financial awareness, but also to provide finer diagnosis of professionally relevant competence.

Several general observations can already be made. First, classical public benchmarks remain an important and useful reference point, but they do not always provide sufficient differentiation among models in more challenging professional settings. Second, exam-oriented benchmarks create a stronger stratification of results and are better at revealing performance degradation as difficulty increases. Third, applied datasets in technical analysis and derivatives complement the picture, because some models that perform well on public benchmarks show substantially weaker results precisely in these specialized domains.

\section{Analysis of Results}
\label{sec:analysis}

\subsection{Public Benchmarks vs.\ FINESSE-Bench: The Transfer Gap}
\label{sec:analysis_transfer_gap}

One of the central questions of our study is how well results on classical open financial benchmarks transfer to the more professionally oriented benchmark groups in FINESSE-Bench. To answer this question, we compare aggregated model results across three task groups: \textit{public benchmarks}, \textit{exam-like}, and \textit{trading/TA}. In Table~\ref{tab:transfer_gap}, for each model we report not only aggregated accuracy values, but also two transfer-gap indicators: $\Delta_{\text{public}\rightarrow\text{exam}}$ and $\Delta_{\text{public}\rightarrow\text{ta}}$, that is, the difference between performance on classical open benchmark datasets and performance on the two FINESSE-Bench groups.

\begin{table}[H]
\footnotesize
\setlength{\tabcolsep}{2.5pt}
\caption{Transfer Gap Between Classical Open Financial Benchmarks and FINESSE-Bench Benchmark Groups}
\label{tab:transfer_gap}
\begin{tabular}{lrrrrr}
\toprule
model & public score & exam-like score & trading/TA score & $\Delta_{\text{public}\rightarrow\text{exam}}$ & $\Delta_{\text{public}\rightarrow\text{ta}}$ \\
\midrule
Claude Sonnet 4.6 & 0.9066 & 0.8763 & 0.8245 & 0.0303 & 0.0821 \\
Claude Sonnet 3.7 & 0.8824 & 0.8261 & 0.7854 & 0.0563 & 0.0970 \\
Kimi K2.5 & 0.9001 & 0.8740 & 0.8498 & 0.0261 & 0.0503 \\
MiMo-V2-Flash & 0.8499 & 0.5880 & 0.6450 & 0.2619 & 0.2049 \\
GPT‑5.4& 0.8880 & 0.7384 & 0.7969 & 0.1496 & 0.0911 \\
GPT‑5.2& 0.8987 & 0.8595 & 0.8424 & 0.0392 & 0.0563 \\
MiniMax M2.5 & 0.8919 & 0.8005 & 0.7623 & 0.0914 & 0.1296 \\
MiniMax M2.1 & 0.8445 & 0.7712 & 0.7462 & 0.0733 & 0.0983 \\
Qwen3.5-Plus-02-15 & 0.8856 & 0.8776 & 0.8475 & 0.0080 & 0.0381 \\
Qwen3.5-Flash-02-23 & 0.8794 & 0.8408 & 0.7894 & 0.0386 & 0.0900 \\
Qwen3.5-397B-A17B & 0.8784 & 0.8443 & 0.8331 & 0.0341 & 0.0453 \\
Qwen3.5-122B-A17B & 0.8794 & 0.8568 & 0.8279 & 0.0226 & 0.0515 \\
Qwen3.5-35B-A3B & 0.8747 & 0.8390 & 0.7883 & 0.0357 & 0.0864 \\
Qwen3.5-27B & 0.8775 & 0.8435 & 0.8038 & 0.0340 & 0.0737 \\
Qwen3.5-9B & 0.8726 & 0.7628 & 0.7284 & 0.1098 & 0.1442 \\
GLM-5 & 0.8724 & 0.8727 & 0.8130 & -0.0003 & 0.0594 \\
GLM-4.7 & 0.8685 & 0.8639 & 0.8205 & 0.0046 & 0.0480 \\
Llama 4 Maverick & 0.8489 & 0.7849 & 0.7451 & 0.0640 & 0.1038 \\
Qwen3-235B-A22B-Thinking-2507  & 0.8673 & 0.8381 & 0.8027 & 0.0292 & 0.0646 \\
Qwen3-32B& 0.8699 & 0.7938 & 0.7479 & 0.0761 & 0.1220 \\
Qwen3-14B& 0.8687 & 0.7512 & 0.7019 & 0.1175 & 0.1668 \\
Qwen3-8B& 0.8506 & 0.6891 & 0.6744 & 0.1615 & 0.1762 \\
DeepSeek-V3.2 & 0.8571 & 0.7379 & 0.7601 & 0.1192 & 0.0970 \\
DeepSeek-R1-0528 & 0.8610 & 0.8266 & 0.7848 & 0.0344 & 0.0762 \\
Mistral-Small-3.2-24B-Instruct-2506 & 0.8564 & 0.5751 & 0.5892 & 0.2813 & 0.2672 \\
Fino1-8B & 0.7974 & 0.4661 & 0.4372 & 0.3313 & 0.3602 \\
Fin-R1 & 0.8569 & 0.5650 & 0.5212 & 0.2919 & 0.3357 \\
Fin-o1-8B & 0.8538 & 0.5783 & 0.5495 & 0.2755 & 0.3043 \\
GigaChat3-10B-A1.8B-bf16 & 0.8087 & 0.4688 & 0.4971 & 0.3399 & 0.3116 \\
YandexGPT Pro 5.1 & 0.8438 & 0.6426 & 0.6289 & 0.2012 & 0.2149 \\
T-pro-it-2.0 & 0.8689 & 0.7796 & 0.7255 & 0.0893 & 0.1434 \\
\bottomrule
\end{tabular}
\end{table}
\FloatBarrier

The results show that, for almost all models considered, performance on the FINESSE-Bench groups is lower than on classical open financial benchmark datasets. In other words, most models exhibit a positive transfer gap in both directions: \textit{public benchmarks} $\rightarrow$ \textit{exam-like} and \textit{public benchmarks} $\rightarrow$ \textit{trading/TA}. This alone suggests that strong performance on standard public benchmark datasets does not guarantee equally strong performance on more professionally oriented financial tasks.

A small set of strong models appears comparatively robust. In particular, Qwen3.5-Plus-02-15 shows the smallest gap with respect to the \textit{exam-like} group ($0.0080$) and one of the smallest gaps with respect to \textit{trading/TA} ($0.0381$). A similar profile is observed for GLM-5 and GLM-4.7: for GLM-5, the gap on the \textit{exam-like} group even becomes slightly negative ($-0.0003$), meaning that the aggregated result on exam-like tasks is marginally higher than on the public benchmark sets. Among frontier models, Claude Sonnet 4.6, GPT-5.2, and Kimi K2.5 also maintain comparatively small gaps.

By contrast, a number of models exhibit a pronounced gap between results on classical open benchmarks and results on FINESSE-Bench. This is especially visible for small specialized financial models and for some smaller or weaker general-purpose models. For example, Fino1-8B, Fin-R1-7B, and Fin-o1-8B show gaps on the order of $0.27$--$0.36$, while Mistral-Small-3.2-24B-Instruct-2506, GigaChat3-10B-A1.8B-bf16, and YandexGPT Pro 5.1 also display very substantial gaps. These observations are particularly important because they show that a model may appear competitive on classical open financial benchmarks while still suffering a marked performance drop on tasks that are closer to professional exam-style and applied scenarios.

\begin{figure}[!htbp]
    \centering
    \includegraphics[width=0.72\textwidth]{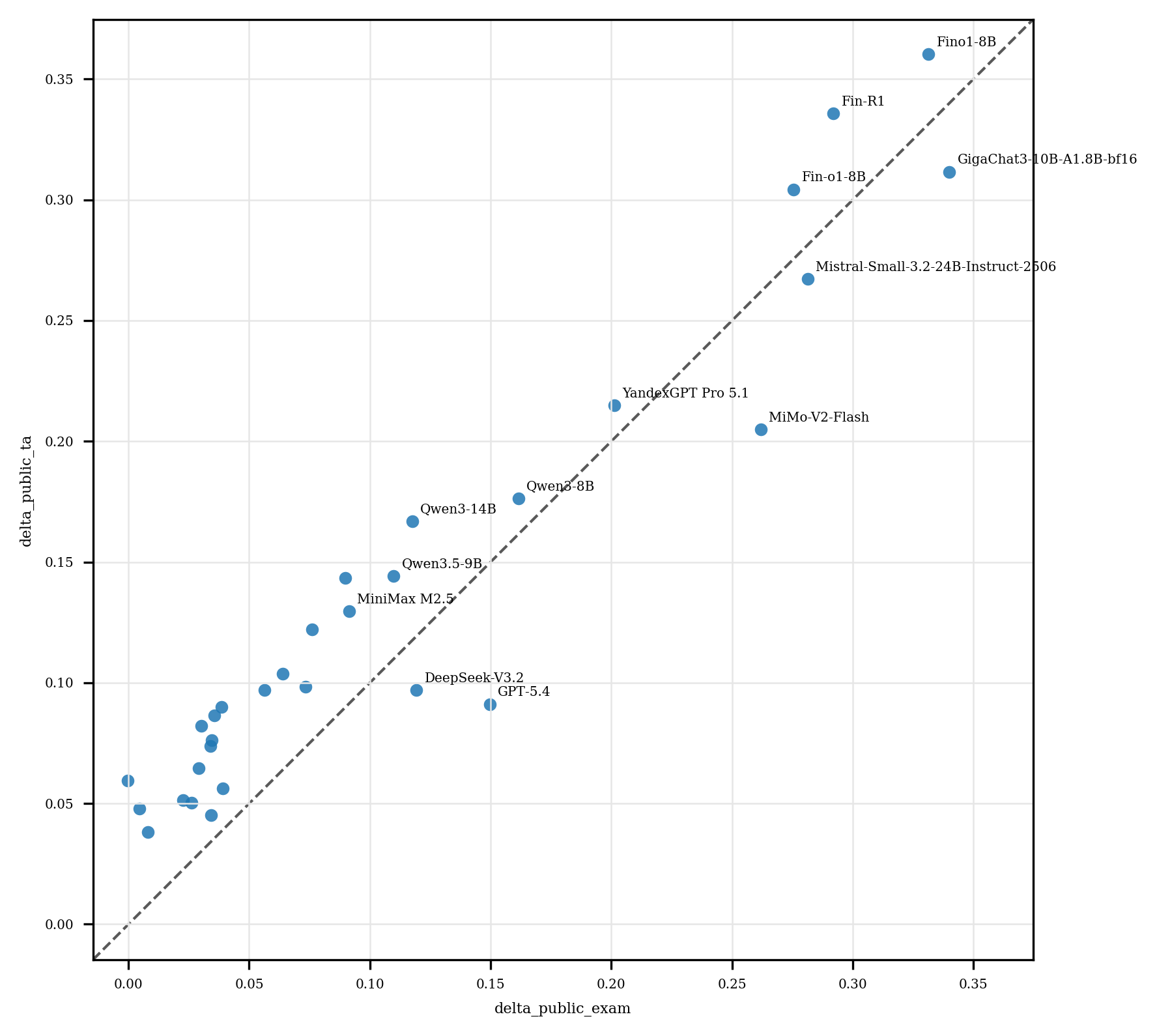}
    \caption{Comparison of transfer gaps from classical open financial benchmarks to FINESSE-Bench benchmark groups. The X-axis shows the gap between \textit{public benchmarks} and \textit{exam-like}, while the Y-axis shows the gap between \textit{public benchmarks} and \textit{trading/TA}. The dashed line corresponds to equal degradation on both benchmark groups.}
    \label{fig:transfer_gap_scatter}
\end{figure}

Additional intuition is provided by Figure~\ref{fig:transfer_gap_scatter}, where the X-axis shows $\Delta_{\text{public}\rightarrow\text{exam}}$ and the Y-axis shows $\Delta_{\text{public}\rightarrow\text{ta}}$. This visualization emphasizes not the absolute level of quality, but the pattern of transfer between benchmark groups. Most points lie in the upper-right region of the plot, corresponding to simultaneous degradation on both FINESSE-Bench groups relative to \textit{public benchmarks}. A substantial fraction of models also lies near the diagonal, indicating comparable declines on exam-like and trading/TA tasks. However, several models deviate from this line, exhibiting asymmetric transfer profiles.

\FloatBarrier

Substantively, this result supports one of the key motivations behind FINESSE-Bench. Classical open benchmark datasets remain an important reference point for evaluating financial language models, but by themselves they do not always reflect how well performance transfers to more professionally oriented financial tasks. FINESSE-Bench, in turn, makes such discrepancies visible. In other words, it does not replace existing public benchmarks; rather, it complements them by exposing the gap between ``good performance on a familiar public format'' and ``more robust financial competence'' on exam-oriented and applied benchmark groups.

\subsection{Hierarchy of Difficulty: CFA-like Level~1 \texorpdfstring{$\rightarrow$}{->} CFA-like Level~2 \texorpdfstring{$\rightarrow$}{->} CFA-like Level~3}
\label{sec:analysis_cfa_hierarchy}

One of the key design principles of FINESSE-Bench is explicit difficulty hierarchy. Unlike benchmark sets in which all questions form a relatively homogeneous mixture, FINESSE-Bench includes a sequence of CFA-like levels that makes it possible to evaluate how model performance changes when moving from foundational tasks to more challenging analytical and expert scenarios. To this end, we separately consider three subsets: CFA-like Level~1, CFA-like Level~2, and CFA-like Level~3.

Table~\ref{tab:cfa_hierarchy} reports results for selected models together with three degradation indicators: $\Delta_{L1\rightarrow L2}$, $\Delta_{L2\rightarrow L3}$, and $\Delta_{L1\rightarrow L3}$, i.e., the differences in performance between adjacent levels and between the two ends of the hierarchy.
\begin{table}[!htbp]
\centering
\small
\caption{Performance Degradation Across the CFA-like Difficulty Hierarchy: Level~1 $\rightarrow$ Level~2 $\rightarrow$ Level~3}
\label{tab:cfa_hierarchy}
\begin{tabular}{lrrrrrr}
\toprule
model & Level 1 & Level 2 & Level 3 & $\Delta_{L1\rightarrow L2}$ & $\Delta_{L2\rightarrow L3}$ & $\Delta_{L1\rightarrow L3}$ \\
\midrule
Claude Sonnet 4.6 & 0.8979  & 0.9181  & 0.8239  & -0.0202  & 0.0942  & 0.0740  \\
Kimi K2.5 & 0.8924  & 0.9181  & 0.8113  & -0.0257  & 0.1068  & 0.0811  \\
GPT‑5.4& 0.7315  & 0.6962  & 0.6509  & 0.0353  & 0.0453  & 0.0806  \\
GPT‑5.2& 0.8736  & 0.8874  & 0.8019  & -0.0138  & 0.0855  & 0.0717  \\
MiniMax M2.5 & 0.8148  & 0.8157  & 0.7170  & -0.0009  & 0.0987  & 0.0978  \\
Qwen3.5-397B-A17B & 0.8756  & 0.8225  & 0.7201  & 0.0531  & 0.1024  & 0.1555  \\
GLM-5 & 0.8859  & 0.8976  & 0.8396  & -0.0117  & 0.0580  & 0.0463  \\
Llama 4 Maverick & 0.8241  & 0.7645  & 0.7170  & 0.0596  & 0.0475  & 0.1071  \\
Qwen3-235B-A22B-Thinking-2507  & 0.8700  & 0.8635  & 0.7484  & 0.0065  & 0.1151  & 0.1216  \\
DeepSeek-V3.2 & 0.7961  & 0.6587  & 0.5786  & 0.1374  & 0.0801  & 0.2175  \\
GigaChat3-10B-A1.8B-bf16 & 0.4911  & 0.3345 & 0.4151  & 0.1566  & -0.0806  & 0.0760  \\
YandexGPT Pro 5.1 & 0.6791  & 0.5358  & 0.6006  & 0.1433  & -0.0648  & 0.0785  \\
\bottomrule
\end{tabular}
\end{table}
\FloatBarrier

The observed pattern confirms that the CFA-like hierarchy indeed reflects increasing difficulty overall, but not in the form of strictly monotonic degradation at every adjacent step. Rather, it does so in a more realistic way: deterioration from Level~1 to Level~2 is observed in most, but not all, cases, whereas degradation from Level~1 to Level~3 is present for all models considered without exception. In other words, CFA-like Level~3 consistently emerges as a more difficult slice than Level~1, even if some models occasionally perform on Level~2 comparably to, or slightly better than, Level~1.

This effect is clearly visible for strong frontier or near-frontier models. For example, Claude Sonnet 4.6, Kimi K2.5, GPT-5.2, and GLM-5 show very similar values on Level~1 and Level~2. Yet all of these models lose performance when moving to Level~3. This suggests that Level~2 is not simply a ``harder Level~1,'' but in some cases may align better with particular model strengths, whereas Level~3 imposes more stable requirements for complex synthesis and strategic financial reasoning.

For some models, hierarchical degradation is even stronger and more monotonic. GPT-5.4, Llama 4 Maverick, DeepSeek-V3.2, all show a sequential decline from Level~1 to Level~2 and then to Level~3. The effect is particularly pronounced for DeepSeek-V3.2, where the total gap between the two extremes reaches $0.2175$. Cases like this illustrate especially well that the CFA-like hierarchy can indeed be used as a tool for measuring how well a model retains financial competence as difficulty increases.

The table also shows that local non-monotonicity between adjacent levels is not itself a flaw of benchmark design. On the contrary, it indicates that different levels test not only ``more of the same difficulty,'' but also somewhat different competence profiles. For example, negative values of $\Delta_{L1\rightarrow L2}$ for Claude Sonnet 4.6, Kimi K2.5, GPT-5.2, MiniMax M2.5, and GLM-5 mean that, under the specific task composition of Level~2, these models perform no worse than, and sometimes even slightly better than, on Level~1. However, none of these models avoids degradation on Level~3 relative to Level~1. This is precisely what makes the comparison between Level~1 and Level~3 the most reliable indicator of hierarchical degradation.

Substantively, this result matters for two reasons. First, it confirms that FINESSE-Bench is not merely a collection of heterogeneous exam-style questions, but a meaningful ladder of difficulty. Second, it shows that evaluation based only on basic-level questions may underestimate a model's actual limitations in more advanced financial scenarios. Thus, the CFA-like hierarchy within FINESSE-Bench serves the diagnostic function for which it was originally introduced: it measures not only ``average financial literacy,'' but also the robustness of model quality when moving to more difficult professional tasks.

\subsection{Within-Family Scaling: The Qwen Case}
\label{sec:analysis_qwen_scaling}

An additional way to test the discriminative power of FINESSE-Bench is to examine not only differences between models from different families, but also how well benchmark groups separate models within a closely related family. To this end, we consider the Qwen3 model line: \texttt{Qwen3-8B}, \texttt{Qwen3-14B}, \texttt{Qwen3-32B}, and \texttt{Qwen3-235B-A22B-Thinking-2507}. This slice is particularly useful because the models are architecturally similar, making differences easier to interpret than in comparisons across laboratories and unrelated architectures.

\begin{figure}[!htbp]
    \centering
    \includegraphics[width=0.88\textwidth]{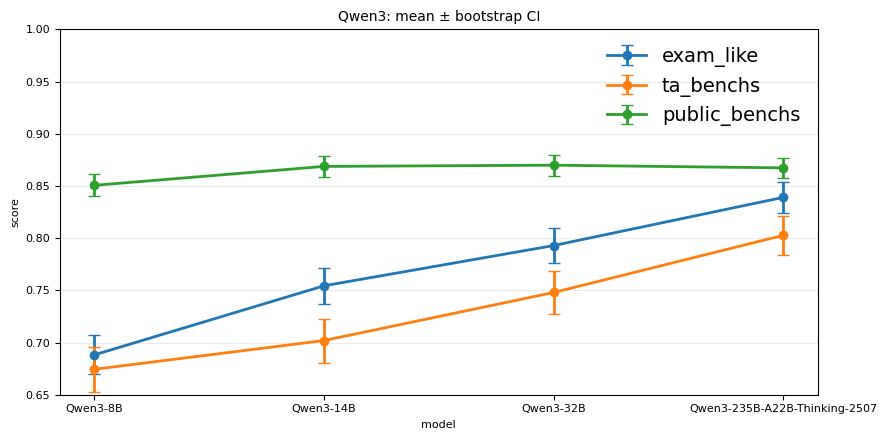}
    \caption{Within-family scaling results for reasoning-oriented models from the Qwen3 family. The figure shows aggregated results for the benchmark groups \textit{public benchmarks}, \textit{exam-like}, and \textit{trading/TA}; vertical bars denote bootstrap confidence intervals.}
    \label{fig:qwen_family_scaling}
\end{figure}
\FloatBarrier

Figure~\ref{fig:qwen_family_scaling} shows three aggregated curves corresponding to the benchmark groups \textit{public benchmarks}, \textit{exam-like}, and \textit{trading/TA}. The observed pattern illustrates one of the key properties of FINESSE-Bench well. On the \textit{public benchmarks} group, results for reasoning-oriented Qwen3 models differ only slightly, remaining tightly clustered in the $0.8506$–$0.8699$ range. In other words, on public open benchmark datasets, the entire family appears relatively compressed, and the gains of stronger models over smaller versions are limited.

A very different picture emerges on the FINESSE-Bench groups. On \textit{exam-like} tasks, performance increases much more substantially, from $0.6891$ for \texttt{Qwen3-8B} to $0.8381$ for \texttt{Qwen3-235B- A22B-Thinking-2507}. The effect is analogous on the \textit{trading/TA} group, where scores rise from $0.6744$ to $0.8027$. Thus, as one moves from smaller reasoning-oriented models to larger ones, FINESSE-Bench captures a much more pronounced improvement in quality than is visible from \textit{public benchmarks} alone. In other words, within this family, FINESSE-Bench provides a clearer scale of distinctions between models.

This observation is important for two reasons. First, it confirms that FINESSE-Bench groups have stronger discriminative power not only in comparisons across families, but also within a single model line. Second, it shows that public benchmark sets in this case partly ``smooth out'' differences between model versions, whereas \textit{exam-like} and \textit{trading/TA} tasks make it easier to see how financial domain competence changes with model scale.

Importantly, the gains on FINESSE-Bench groups do not appear accidental, but substantively coherent. For the Qwen3 family, quality improves when moving from \texttt{8B} to \texttt{14B}, then to \texttt{32B}, and finally to \texttt{235B}, both on exam-like and on trading/TA tasks, in an almost monotonic fashion. This is especially useful for benchmark practice: it shows that FINESSE-Bench can serve not only as a tool for ranking substantially different models, but also as a means of more fine-grained diagnosis of scaling behavior within a single family.

Taken together, this case study reinforces the main thesis of the paper. If the Qwen3 models differ only modestly on classical open financial benchmarks, FINESSE-Bench groups reveal a more pronounced and interpretable gradient of quality. Consequently, FINESSE-Bench is better suited to settings in which it is important to measure not just absolute score, but also the ability of a benchmark to distinguish between models that are similar in architecture and origin.

\subsection{Discriminative Power and Benchmark Saturation}
\label{sec:analysis_saturation}

Beyond metric comparisons, it is important to understand how well individual benchmark sets actually distinguish between models. From this perspective, it is useful to examine not only final scores, but also the structure of question difficulty. If too large a share of questions is solved either by all models or by none, the benchmark has lower discriminative power. By contrast, the most informative questions are those that are answered correctly by a substantial, but not overwhelming, fraction of models. In this work, to approximate this property we use three aggregated characteristics: the share of questions that no model solves (\textit{unanimous fail}), the share of questions that all models solve (\textit{unanimous success}), and the share of questions falling into an intermediate discriminative band, where between 10\% and 90\% of models answer correctly (\textit{mid-band 10--90}).
\begin{table}[!htbp]
\centering
\small
\setlength{\tabcolsep}{2.5pt}
\caption{Saturation and Discriminative-Power Profile of FINESSE-Bench and Public Financial Benchmark Sets}
\label{tab:benchmark_saturation}
\begin{tabular}{lrrrrr}
\toprule
benchmark & $n_{\text{models}}$ & $n_{\text{questions}}$ & unanimous fail(\%) & unanimous success(\%) & mid-band 10--90(\%) \\
\midrule
FinQA & 53    & 1147    & 9.33   & 22.41   & 24.15   \\
ConvFinQA & 53    & 1488    & 3.09   & 34.61   & 18.41   \\
TAT-QA & 52    & 1668    & 1.02   & 45.74   & 12.35   \\
CFA-like Level~1 & 56    & 1069    & 0.94   & 13.10   & 56.97   \\
CFA-like Level~2 & 58    & 293    & 1.71   & 1.71   & 81.23   \\
CFA-like Level~3 & 57    & 318    & 0.00    & 0.94   & 81.13   \\
CMT-like Level~2 & 57    & 251    & 1.59   & 17.53   & 48.21   \\
VLigaBench-ru & 55    & 324    & 4.63   & 4.94   & 66.98   \\
Trading\_TA & 49    & 413    & 4.84   & 21.31   & 44.55   \\
Trading\_derivatives & 50    & 544    & 2.76   & 5.33   & 70.40   \\
CFTe-like Level~1 & 43    & 781    & 0.90   & 21.13   & 46.61   \\
\bottomrule
\end{tabular}
\end{table}
\FloatBarrier

Table~\ref{tab:benchmark_saturation} shows that many FINESSE-Bench components have a substantially higher share of \textit{mid-band 10--90} questions than classical open financial benchmark resources. This measure is especially important for discriminating among models: if a question is solved neither by everyone nor by no one, but only by a subset of models, it is more useful for ranking and comparison. The effect is most pronounced for CFA-like Level~2 ($81.23\%$), CFA-like Level~3 ($81.13\%$), Trading\_derivatives ($70.40\%$), and VLigaBench-ru ($66.98\%$). This means that a large share of questions in these benchmark sets lies in the most informative difficulty region, where models truly separate by quality.

By comparison, on classical open benchmark sets the share of mid-band questions is much lower: $24.15\%$ for FinQA, $18.41\%$ for ConvFinQA, and only $12.35\%$ for TAT-QA. At the same time, these benchmarks have a substantially higher share of questions solved by all models, especially TAT-QA ($45.74\%$) and ConvFinQA ($34.61\%$). This structure does not make public benchmark sets ``bad''; on the contrary, they remain important and useful resources for model comparison. However, the table suggests that, from the standpoint of distinguishing contemporary models, their saturation profile is often less favorable than that of the more specialized FINESSE-Bench groups.

Two extreme cases within FINESSE-Bench are especially illustrative. On the one hand, CFA-like Level~2 and CFA-like Level~3 contain almost no fully saturated questions: the share of \textit{unanimous success} is only $1.71\%$ and $0.94\%$, respectively, while the share of \textit{unanimous fail} also remains very low. In other words, these benchmark sets avoid both extremes: they are neither too easy nor too ``impossible.'' Instead, the bulk of their questions lies in the discriminative difficulty band. On the other hand, CFA-like Level~1 and CMT-like Level~2 show a more mixed but still strong profile: they have a larger share of fully solvable questions, yet the mid-band remains substantial ($56.97\%$ and $48.21\%$, respectively), which makes them useful both for general evaluation and for model ranking.

The profile of the applied benchmark sets is also interesting. Trading\_derivatives exhibits one of the strongest discriminative configurations: a low \textit{unanimous fail} rate ($2.76\%$), a moderate \textit{unanimous success} rate ($5.33\%$), and a very high \textit{mid-band 10--90} share ($70.40\%$). This is consistent with the general motivation of the paper: applied financial tasks in derivatives are capable of revealing differences between models that are not always visible on more classical open benchmark sets. Trading\_TA and CFTe-like Level~1 also appear informative, though to a lesser extent, which matches their more introductory and practice-oriented character.

Taken together, this analysis supports an important claim of the paper: FINESSE-Bench differs not only in thematic breadth and difficulty hierarchy, but also in a more favorable discriminative profile of question difficulty. In other words, the suite is useful not merely because it contains ``different financial topics,'' but because many of its subsets place questions in the difficulty range where contemporary models actually begin to separate from one another. This property makes FINESSE-Bench especially suitable for diagnostic evaluation, for comparing closely matched systems, and for more sensitive measurement of progress in financial LLMs.

\subsection{Group Leaders and Model Profiles}
\label{sec:analysis_leaders_profiles}

Another useful lens on the results of FINESSE-Bench is to examine not only transfer gaps and benchmark discriminability, but also which models lead in different benchmark groups and which models exhibit the most balanced performance profile. To this end, we separately analyze, first, the top-performing models in each aggregated benchmark group and, second, the models with the most stable results simultaneously across \textit{public benchmarks}, \textit{exam-like}, and \textit{trading/TA}.
\begin{table}[!htbp]
\centering
\small
\setlength{\tabcolsep}{2.5pt}
\caption{Top-3 Models by Aggregated Benchmark Group}
\label{tab:group_leaders}
\begin{tabular}{lllllll}
\toprule
benchmark group & rank 1 & score & rank 2 & score & rank 3 & score \\
\midrule
public benchmarks & Claude Sonnet 4.6 & 0.9066  & Kimi K2.5 & 0.9001  & GPT‑5.2& 0.8987  \\
exam-like & Qwen3.5-Plus-02-15 & 0.8776  & Claude Sonnet 4.6 & 0.8763 & Kimi K2.5 & 0.8740  \\
trading/TA & Kimi K2.5 & 0.8498  & Qwen3.5-Plus-02-15 & 0.8475  & GPT‑5.2& 0.8424  \\
\bottomrule
\end{tabular}
\end{table}
\FloatBarrier

Table~\ref{tab:group_leaders} shows that there is no single universal leader across all benchmark groups. On classical open benchmarks, the top position is held by Claude Sonnet 4.6, followed by Kimi K2.5 and GPT-5.2. On exam-like tasks, the leader changes: Qwen3.5-Plus-02-15 ranks first, followed by Claude Sonnet 4.6 and Kimi K2.5. Finally, on the trading/TA group, Kimi K2.5 ranks first, while Qwen3.5-Plus-02-15 and GPT-5.2 take second and third place, respectively. This simple slice is already informative in itself: it shows that the aggregated benchmark groups measure different aspects of financial competence and induce partial reordering of the leaderboard.

It is particularly noteworthy that the Qwen3.5 family appears consistently among the leaders on professionally oriented benchmark groups. In particular, Qwen3.5-Plus-02-15 ranks in the top two on both \textit{exam-like} and \textit{trading/TA} tasks. This aligns well with earlier observations in the paper: Qwen3.5-family models generally exhibit a strong and relatively robust profile in the financial domain, especially on benchmark sets that require not only handling familiar question-answering formats over financial reporting, but also broader subject-matter competence.

However, top-3 rankings by benchmark group alone are insufficient to understand which models are most ``balanced'' in a broader sense. A high rank in one benchmark group does not imply equally strong behavior on the others. It is therefore additionally useful to consider models that achieve a high average result across the three benchmark groups while also maintaining a comparatively strong lower bound on performance.

\begin{table}[H]
\centering
\small
\caption{Most Balanced Models Across the Three Benchmark Groups}
\label{tab:balanced_models}
\begin{tabular}{lrrrrrr}
\toprule
model & public benchmarks & exam-like & trading/TA & mean & min & std \\
\midrule
Kimi K2.5 & 0.9001  & 0.8740  & 0.8498  & 0.8746 & 0.8498 & 0.0252 \\
Qwen3.5-Plus-02-15 & 0.8856  & 0.8776  & 0.8475  & 0.8702 & 0.8475 & 0.0201 \\
Claude Sonnet 4.6 & 0.9066  & 0.8763  & 0.8245  & 0.8691 & 0.8245 & 0.0415 \\
GPT‑5.2& 0.8987  & 0.8595  & 0.8424  & 0.8669 & 0.8424 & 0.0289 \\
Qwen3.5-122B-A17B & 0.8794  & 0.8568  & 0.8279  & 0.8547 & 0.8279 & 0.0258 \\
GLM-5 & 0.8724  & 0.8727  & 0.8130  & 0.8527 & 0.8130 & 0.0344 \\
Qwen3.5-397B-A17B & 0.8784  & 0.8443  & 0.8331  & 0.8519 & 0.8331 & 0.0236 \\
GLM-4.7 & 0.8685  & 0.8639  & 0.8205  & 0.8510 & 0.8205 & 0.0265 \\
Qwen3.5-27B & 0.8775  & 0.8435  & 0.8038  & 0.8416 & 0.8038 & 0.0369 \\
\bottomrule
\end{tabular}
\end{table}
\FloatBarrier

Table~\ref{tab:balanced_models} shows that, in terms of average performance across the three benchmark groups, the most balanced models are Kimi K2.5 and Qwen3.5-Plus-02-15. The profile of Qwen3.5-Plus-02-15 is especially notable: while its average score is slightly below that of Kimi K2.5, it exhibits the smallest standard deviation among the top entries ($0.0201$), making it one of the most even and robust participants in the comparison. Kimi K2.5, in turn, combines a high average score ($0.8746$) with a strong lower bound ($0.8498$), making it an especially strong balanced model within our benchmark groups.

Against this background, the distinction between ``peak'' leadership and ``balanced'' financial competence becomes clear. For example, Claude Sonnet 4.6 ranks first on the \textit{public benchmarks} group, but its minimum score across the three benchmark groups ($0.8245$) is lower than that of Kimi K2.5 and Qwen3.5-Plus-02-15. Similarly, GLM-5 is very strong on \textit{exam-like} tasks, but its profile across the three benchmark groups is less even than that of the leaders in average robustness. This observation is important in practical terms: model selection depends not only on the highest metric value on one benchmark group, but also on how consistently the model behaves across different financial domains.

Taken together, this analysis shows that FINESSE-Bench is useful not only for constructing a leaderboard, but also for developing a finer typology of models. Some systems emerge as ``local leaders'' on particular benchmark groups, while others exhibit a more uniform profile across different subject areas. Such distinctions are difficult to observe on any single benchmark, but become visible precisely in a benchmark suite where professional financial competence is decomposed into several substantively different directions.

\section{Practical Implications and Limitations}
\label{sec:practical_limitations}

\subsection{Practical Implications for Model Development and Selection}

One practical takeaway from our work is that a benchmark suite in finance should be judged not only by breadth of thematic coverage, but also by its usefulness within the real model-development cycle. In this respect, FINESSE-Bench offers several applied advantages.

First, a substantial portion of FINESSE-Bench is presented in multiple-choice format (MCQ). This format is widely used in knowledge and reasoning evaluation for language models, including MMLU-like setups \cite{hendrycks2021mmlu, li2024cmmlu}. For fine-tuning practice and intermediate validation, this is particularly important: MCQ tasks are easy to integrate into rapid evaluation loops, allow reproducible checkpoint comparison, and, where needed, support logit-based evaluation without requiring complex post-processing of free-form outputs.

Second, FINESSE-Bench is useful as a tool for \textit{stage-wise diagnosis}. Our results show that a model may appear competitive on classical open financial benchmarks, yet still suffer a substantial drop on the \textit{exam-like} or \textit{trading/TA} groups. In practical development, this means that relying on only one popular benchmark may lead to overestimation of a model's true domain competence. A benchmark suite organized by groups and difficulty levels is better suited for intermediate model selection and for tracking which financial skills are improving---or, conversely, remain weak---during fine-tuning.

Third, FINESSE-Bench makes it possible to distinguish not only ``local leaders'' but also \textit{balanced models}. As shown in the benchmark-group analysis, some systems achieve very strong results on one group of tasks but exhibit less stable behavior on others. For applied financial scenarios, this distinction may be critical: when selecting a model for a practical use case, a consistently strong profile across several financial subdomains may matter more than a peak score on a single benchmark.

\subsection{Limitations of the MCQ Format and Interpretation of Benchmark Results}

Despite its practical advantages, the MCQ format should not be overinterpreted as an ideal substitute for real professional reasoning. Recent work suggests that multiple-choice questions may partially simplify the task for models through option structure, elimination heuristics, and other artifacts that are not always equivalent to genuine free-form reasoning \cite{rethinkingmc2025}. Therefore, the substantial MCQ component of FINESSE-Bench should be regarded as an engineering and evaluation advantage, but not as a guarantee of full equivalence to real professional tasks.

For this reason, FINESSE-Bench is not limited to MCQ alone: it also includes numerical-answer and short-answer tasks, and some benchmarks are structured around more applied scenarios and linked cases. We view this combination as a compromise between reproducibility of evaluation, usability during model development, and the need to cover more realistic forms of financial reasoning.

\subsection{Limitations of Data Provenance and Domain Coverage}

Despite the strong diagnostic profile of FINESSE-Bench, our work has several limitations related to data provenance. Questions were collected from public internet sources, educational materials, training tasks, and publicly available preparation formats, but the provenance of individual items was not documented completely. This limits full traceability of all benchmark elements and necessitates a cautious release policy. For this reason, we release FINESSE-Bench for non-commercial research use and maintain a removal mechanism for disputed materials through the project repository.

In addition, as with other open benchmarks, the risk of partial contamination cannot be completely ruled out: some questions may have appeared in the training data of certain models. Such risks are also discussed in the financial benchmark literature, where fidelity and benchmark design quality are emphasized alongside breadth of coverage \cite{tang2025financereasoning}. We do not present FINESSE-Bench as a benchmark fully protected against contamination, but rather as a more substantive and more diagnostic step toward evaluation of professional financial competence.

Another limitation concerns domain coverage. Although FINESSE-Bench spans a broad range of financial topics---from financial reporting and corporate finance to technical analysis, derivatives trading, and Russian-language olympiad problems---it does not claim to cover the entire financial industry. The current version does not include full benchmark blocks for areas such as regulatory compliance, banking risk modeling, insurance analytics, corporate liquidity management, or financial law. In this sense, FINESSE-Bench should be viewed as an extensible benchmark suite rather than a final standard for all financial competencies.

\section{Conclusion}
\label{sec:conclusion}

In this work, we introduced FINESSE-Bench, a hierarchical suite of eight specialized benchmarks for evaluating large language models in finance. Unlike benchmarks focused primarily on question answering over financial reporting, or broader but less structured collections of financial tasks, FINESSE-Bench emphasizes the combination of three properties: difficulty hierarchy, breadth of domain coverage, and professionally oriented applied scenarios.

Our analysis shows that FINESSE-Bench indeed contributes new diagnostic information beyond classical open financial benchmarks. First, we observe a consistent transfer gap: strong results on public benchmarks do not always carry over to FINESSE-Bench task groups, especially \textit{exam-like} and \textit{trading/TA}. Second, the CFA-like hierarchy within FINESSE-Bench makes it possible to capture performance degradation as difficulty increases and thereby measure not only a model's foundational financial literacy, but also the robustness of its behavior on more advanced tasks. Third, the saturation and discriminative-power analysis shows that many FINESSE-Bench subsets place a substantial share of questions in the most informative difficulty zone, where contemporary models genuinely begin to diverge in performance. Finally, the within-family Qwen comparison and the analysis of group leaders show that FINESSE-Bench is useful both for cross-family comparison and for more fine-grained evaluation of scaling behavior within closely related model lines.

These findings support a broader conclusion: evaluating financial LLMs requires more than relying solely on popular open benchmarks with established formats. Such resources remain important and useful, but a more complete assessment of domain-specific financial competence requires a benchmark suite that simultaneously tests transferability, difficulty hierarchy, domain specialization, and robustness across distinct subject groups. This is precisely the role that FINESSE-Bench is intended to serve.

We view the current version of FINESSE-Bench as a first step rather than a completed standard. Natural directions for future work include expanding coverage of financial subdomains, strengthening the multilingual component, increasing the share of more open-ended tasks, and conducting further expert validation of judge-model-based scoring. Nevertheless, even in its current form, FINESSE-Bench provides a useful instrument for model comparison, intermediate validation in fine-tuning pipelines, and more substantive diagnosis of the strengths and weaknesses of modern LLMs in finance.

\bibliographystyle{unsrtnat}
\bibliography{references_en}

\end{document}